\documentclass[runningheads]{llncs}
\let\paragraph\oldparagraph
\let\subparagraph\oldsubparagraph
\usepackage{times}
\usepackage{epsfig}
\usepackage{graphicx}
\usepackage{mathrsfs}
\usepackage{amssymb}
\usepackage{amsmath} 
\usepackage{mathtools}
\usepackage{multirow}
\usepackage{multicol}
\usepackage{makecell}
\usepackage{caption}
\usepackage{footmisc}
\usepackage[compact]{titlesec}

\usepackage{hyperref}
\usepackage{arydshln} 
\usepackage{bbm}
\usepackage{xcolor}

\usepackage{algorithm}
\usepackage[noend]{algorithmic}

\usepackage{marvosym}

\usepackage[resetlabels,labeled]{multibib}
\newcites{R}{Appendix Reference}

\hypersetup{
    colorlinks=true,
    linkcolor=blue,
    filecolor=magenta,      
    urlcolor=cyan,
}

\DeclareMathOperator*{\argmin}{arg\,min}

\begin{document}

\title{PGADA: Perturbation-Guided Adversarial Alignment for Few-shot Learning Under the Support-Query Shift}

\titlerunning{PGADA: Perturbation-Guided Adversarial Alignment }

\author{Siyang Jiang\textsuperscript{\dag\ddag}, 
Wei Ding\textsuperscript{\dag}, 
Hsi-Wen Chen\textsuperscript{\dag}, 
Ming-Syan Chen\textsuperscript{\dag \Letter} }

\institute{\dag Department of Electrical Engineering,
National Taiwan University\\
\ddag School of Mathematics and Statistics, Huizhou University\\  
{\tt\small  \{syjiang,wding,hwchen\}@arbor.ee.ntu.edu.tw, mschen@ntu.edu.tw}}

\maketitle          
\begin{abstract}
Few-shot learning methods aim to embed the data to a low-dimensional embedding space and then classify the unseen query data to the seen support set. While these works assume that the support set and the query set lie in the same embedding space, a distribution shift usually occurs between the support set and the query set, i.e., \emph{the Support-Query Shift}, in the real world. Though optimal transportation has shown convincing results in aligning different distributions, we find that the small perturbations in the images would significantly misguide the optimal transportation and thus degrade the model performance. To relieve the misalignment, we first propose a novel adversarial data augmentation method, namely \emph{Perturbation-Guided Adversarial Alignment (PGADA)}, which generates the \emph{hard} examples in a self-supervised manner. In addition, we introduce \emph{Regularized Optimal Transportation} to derive a smooth optimal transportation plan. Extensive experiments on three benchmark datasets manifest that our framework significantly outperforms the eleven state-of-the-art methods on three datasets. Our code is available at \url{https://github.com/772922440/PGADA}.


\end{abstract}

\section{Introduction}
\label{sec:introduction}

\begin{figure*}[t]
  \centering
  \includegraphics[width= \linewidth]{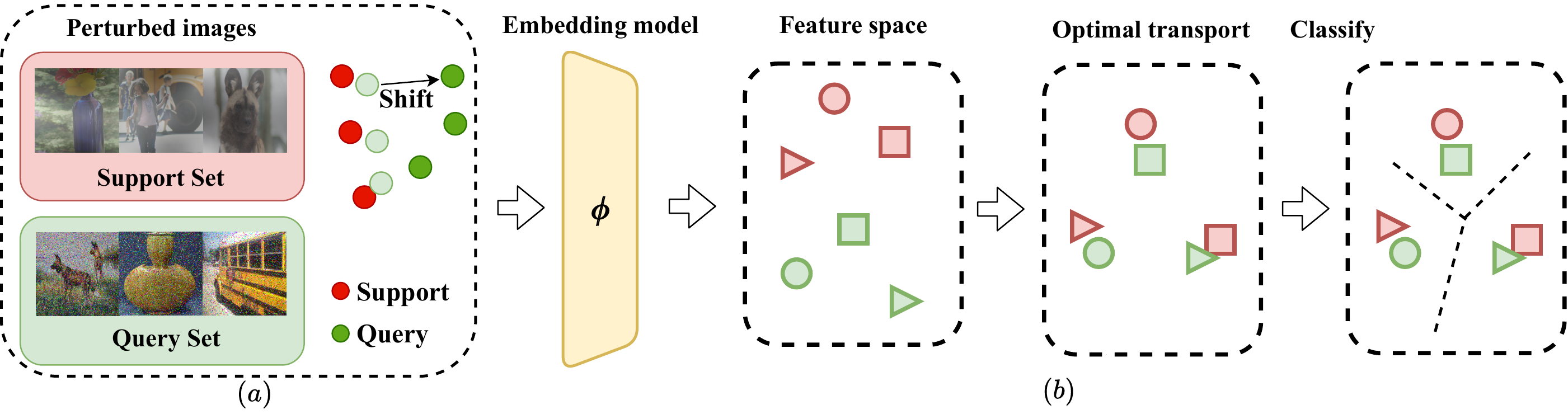}
  \caption{(a) Illustration of the support-query shift in a 3-way 1-shot classification task, where the support set and the query set are embedded into different distributions. (b) First, we embed the support set and the query set via an embedding model $\phi$ into a feature space. Next, optimal transportation is employed to align the support set (red) and the query set (green). But, the small perturbations may misguide the transported results, leading to wrong predictions, i.e., classify the green circle to the red triangle.}
  \label{fig:flow}
\end{figure*}

Recently, deep learning models have celebrated success in several computer vision tasks~\cite{phoo2020self,jiang2019fund,jiang2021dataflow}, which require large amounts of labeled data. However, collecting sufficient labels to train the model parameters involves considerable human effort, which is unacceptable in practice~\cite{garcia2017few}. Moreover, the labels of training data and testing data are usually disjoint, i.e.,  the labels in the testing phase are unseen in the training phase. In contrast, few-shot learning aims to learn the model parameter by a handful of training data (support sets) and adapt to the testing data (query sets)~\cite{vinyals2016matching,finn2017model,bennequin2021bridging}, effectively. In the testing phase, the model classifies the query set into the support sets according to the distance of their embeddings.

While these approaches extract rich contextual features from the images, the embeddings from the support set and the query set usually exhibit certain distribution shifts, i.e., \emph{the Support-Query Shift}~\cite{bennequin2021bridging}. For example, the images are captured by various devices, e.g., smartphones and single-lens reflex cameras in different environments, e.g., foggy and high-luminance. Since the learned embeddings are located in different spaces, degrading the model performance, \emph{optimal transportation}~\cite{courty2016optimal} has been proposed to stage the embeddings from different domains into the same latent space. However, in this paper, we theoretically prove that optimal transport can be easily misguided by small perturbations in the images (as illustrated in Fig.~\ref{fig:flow}). 

Meanwhile, several training techniques~\cite{zhao2020maximum,gong2021maxup} have been proposed to derive a more robust embedding model against the perturbations. On the one hand, data augmentation methods transform a single image~\cite{simonyan2014very} or combine multiple images~\cite{yun2019cutmix,zhang2017mixup} to create more training samples at pixel level. However, these methods cannot create new information which is not included in the given data~\cite{zhao2020maximum}. On the other hand, adversarial training methods such as projected gradient descent (PGD)~\cite{samangouei2018defense}, AugGAN~\cite{huang2018auggan} are used to find the perturbed images to confuse the model, i.e., predicting an incorrect label, as the additional training samples. However, these methods usually require numerous iterations to generate the adversarial examples by optimizing  a predefined adversarial loss, which is computationally intensive. Besides, a clear trade-off has been shown between the accuracy of a classifier and its robustness against adversarial examples~\cite{gong2021maxup}.

To address the above issues, we propose \emph{Perturbation-Guided Adversarial Alignment (PGADA)} to relieve the negative effect caused by the small perturbations in the support-query shift. PGADA aims to generate the perturbed data as the \emph{hard} examples, i.e., less similar to the original data point in the embedding space but still classified into the same class. Next, the model is trained on these generated data by minimizing the empirical risk to enhance the embedding model's robustness of noise tolerance. We further introduce \emph{smooth optimal transport}, which regularizes the negative entropy of the transportation plan to take more query data points as the anchor nodes, leading to a higher error tolerance of the transportation plan.

The contributions of this work are summarized as follows.
\begin{itemize}
     \item We formally investigate how the perturbation in the images would affect the results of optimal transportation under \emph{the Support-Query Shift}. 
    \item We propose \emph{Perturbation-Guided Adversarial Alignment (PGADA)} to relieve the misalignment problem from small perturbations via deriving a more robust feature extractor and a smooth transportation plan under distribution shifts.
    \item Extensive experiments manifest that PGADA outperforms eleven state-of-the-art methods by $10.91\%$ on three public datasets. 
\end{itemize}

\section{Preliminary}
\label{sec:premilinary}
\subsection{Few-shot Learning}
Given a labeled support set
$\mathcal{S} = \cup_{c \in \mathcal{C}} \mathcal{S}^c$, with $\mathcal{C}$ classes, where each class $c$ has $|S^c|$ labeled examples, the goal of few-shot learning is to classify the query set $\mathcal{Q}= \cup_{c \in \mathcal{C}} \mathcal{Q}^c$ into these $\mathcal{C}$ classes. Let $\phi$ denotes the embedding model $\phi(x) \in \mathrm{R}^d $, which encodes the  data point $x$ to the $d$-dimensional feature. $\phi$ is learned from a labeled training set $\mathcal{D} = \{x_i,y_i\}_{i\in[1,|\mathcal{D}|]}$, where $x_i$ is the data point and $y_i$ is the corresponding label. The embedding model can be learned by empirical risk minimization (ERM),
\begin{equation*}
    \min_{\phi,\theta} E_{\{x,y\}\sim \mathcal{D}}[L(\theta(\phi(x)),  y)],
\end{equation*}
where $\theta$ is a trainable parameter to map the embedding $\phi(x_i)$ to the class $y_i$.

Through the embedding model $\phi$, we can encode the data points in support set  (i.e., $x_{s,i} \in \mathcal{S}$) and query set (i.e., $x_{q,j} \in \mathcal{Q}$) to the feature $\phi(x_{s,i})$ and $\phi(x_{q,j})$, respectively. These features are used as input to a comparison function $M$, which measures the distance , e.g., $l_2$-norm, between two samples. Specifically, we classify the query example $\phi(x_{q,j})$ by averaging the embedding $\phi(x^c_{s,i})$ of the support set in class $\mathcal{S}^c$, which can be written as follows.
\begin{equation*}\small
  \phi^c(x_{s}) = \frac{1}{|S^c|} \sum_{x_{s,i} \in S^c} \phi(x_{s,i}), \quad 
  y_q= \argmin_{c} M(\phi^c(x_{s}) ,\phi(x_{q,j})). 
\end{equation*}



\subsection{The Support-Query Shift and Optimal Transportation}
The conventional few-shot learning methods assume the support set and the query set lie in the same distribution. A more realistic setting is that the support set $\mathcal{S}$ and the query set $\mathcal{Q}$ follow different distributions, i.e., the support-query shift~\cite{bennequin2021bridging}. While these two sets are sampled from different distributions $\mu_s$ and $\mu_q$, the embeddings for the support set  $\mathcal{S}$ (i.e., $\phi(x_{s})$) and the query set $\mathcal{Q}$ (i.e., $\phi(x_{q})$) are likely to lie in different embedding spaces. Thus, it would lead to a wrong classification result via the comparison module $M(\phi(x_{s}),\phi(x_{q}))$~\cite{bennequin2021bridging}.

To tackle with the support-query shift, optimal transportation~\cite{courty2016optimal} is one of the effective techniques to align different distributions by a transportation plan $\pi(\mu_s, \mu_q)$, which can formally be written as follows.
\begin{align}\small
    W (\mu_s,\mu_q)=  \inf_{\pi \in \Pi(\mu_s,\mu_q)} \int  w(x_s,x_q)d\pi(x_s,x_q),
\end{align}
where $\Pi(\mu_s, \mu_q)$ is the set of transportation plans (or couplings) and $w$ is the cost function, and $W$ is the overall cost of transporting distribution $\mu_s$ to $\mu_q$. In our practice, we select $l_2$-norm of the embedding vector, i.e., $\Vert \phi(x_{s}) - \phi(x_{q})\Vert^2_2$, as our distance function $w$.

Since there are only finite samples for both the support set $x_{s,i} \in \mathcal{S}$ and the query set $x_{q,j} \in \mathcal{Q}$, the discrete optimal transportation adopts the empirical distributions to estimate the probability mass function $\hat{\mu}_s =\sum \delta_{s,i}$, and $\hat{\mu}_q = \sum \delta_{q,j}$, where $\delta_{s,i}$ and $\delta_{q,j}$ is the Dirac distribution. We obtain
\begin{align}\small
\label{eq:basic_ot}
     & \pi^{\ast} = \argmin_{\pi} \sum_{\substack{x_{s,i} \sim \hat{\mu}_{s}\\ x_{q,j} \sim \hat{\mu}_{q}} } w(x_{s,i},x_{q,j}) \pi(x_{s,i},x_{q,j})
\end{align}
Then, Sinkhorn's algorithm~\cite{cuturi2013sinkhorn} is adopted to solve the optimal transportation plan $\pi^{\ast}$.

Equipped with the optimal plan $\pi^{\ast}$, we transport the embeddings of the support set $\phi(x_{s,i})$ to $\hat{z}_{s,i}$ by barycenter mapping~\cite{courty2016optimal} to adapt the support set to the query set.
\begin{equation}\small
    \hat{\phi}(x_{s,i}) =\frac{\sum_{x_{q,j}\in\mathcal{Q}}\pi^{\ast}(x_{s,i},x_{q,j}) \phi(x_{q,j}) }{ \sum_{x_{q,j}\in\mathcal{Q}}\pi^{\ast}(x_{s,i},x_{q,j})}.
    \label{eq:trans_q}
\end{equation}
$\hat{\phi}(x_{s,i})$ denotes the transported embedding of $x_{s,i}$. Therefore, we can correctly measure the distance metric  $M(\hat{\phi}(x_{s,i}), \phi(x_{q,j}))$ in a shared embedding space.

\section{Methodology}
\label{sec:method}
Here, we first investigate the misestimation of optimal transport of perturbed images. Then, we illustrate our framework, namely \emph{Perturbation-Guided Adversarial Alignment (PGADA)}, which relieves the perturbation in the images and derives a more robust embedding model. In addition, a regularized optimal transportation is introduced to align the support set and the query set better, which takes more data points 
from the query set as anchors to enhance the error tolerance.

\subsection{Motivation}
\label{sec:motivation}
We observe that optimal transportation has a challenge, which comes from the quality of the embedding $\phi(x)$, i.e., the perturbation in the image may misguide the transportation plan. For example, clean images' embeddings may give a better transportation plan than those of foggy images. With some derivation, we formally estimate the error of transported embedding $\hat{\phi}(x_{s,i})$ in Eq.~(\ref{eq:trans_q}) as follows,
    
    


\begin{theorem}
The error of the transported embedding is 
\begin{equation*}\small
        E[\Vert{\hat{\phi}(x_{s,i}) - \hat{\phi}_{\sigma}(x_{s,i})}\Vert_2^2] = \sqrt{d (\sigma_s^2 +\sigma_q^2)}, 
\end{equation*} 
where $\hat{\phi}_{\sigma}(x_{s,i})$ is the transported embedding from the perturbed distribution $W_\sigma(\mu_s,\mu_q)$.  $W_\sigma(\mu_s,\mu_q) \coloneqq W(\mu_s*\mathcal{N}_{\sigma_s},\mu_q*\mathcal{N}_{\sigma_q})$ denotes the original support and query set distributions $\mu_s$ and $\mu_q$ being perturbed with Gaussian noises $\sigma_s$ and $\sigma_q$, and $*$ is the convolution operator.
\label{thm:err}
\end{theorem}

As the noise level, i.e., $\sigma_s$, and $\sigma_q$, increases, it is more likely to mislead the transportation plan and alleviate the model's performance. Therefore, it's non-trivial to learn a better embedding model $\phi$ having a better capability of \emph{noise tolerance} such that $\phi(x_p) \approx \phi(x)$, where $x_p$ is original image $x$ with small perturbation. 

\subsection{Perturbation-Guided Adversarial Alignment (PGADA)}
\begin{figure*}[t]
  \centering
  \includegraphics[width=\linewidth]{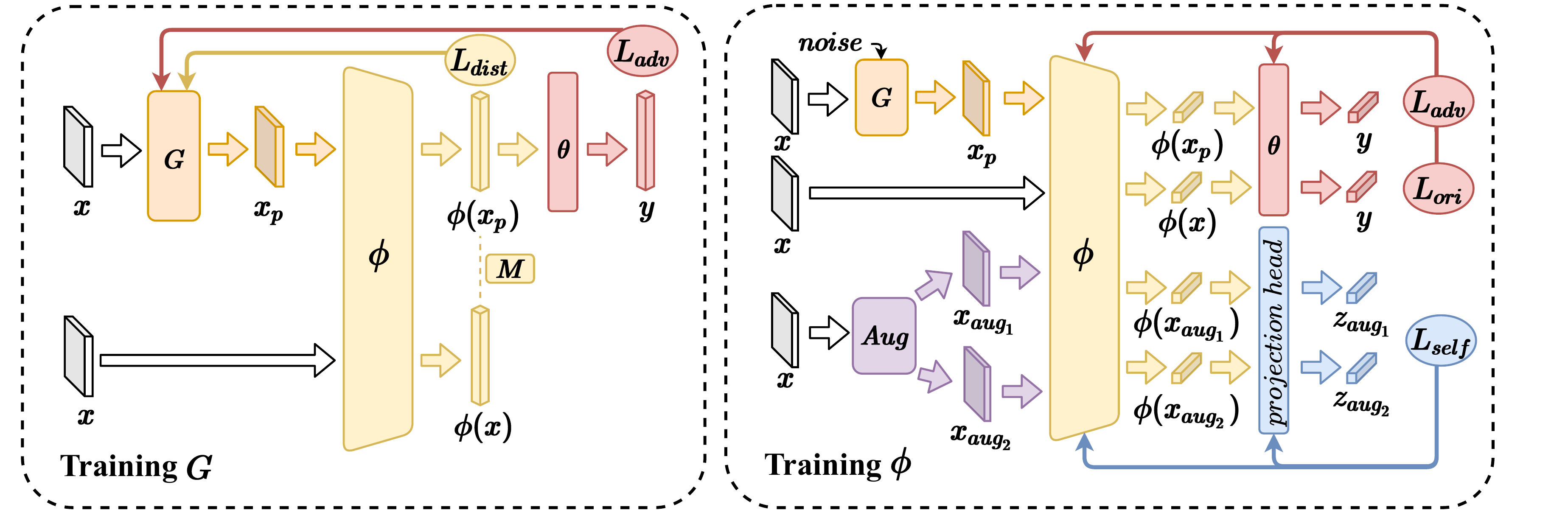}
  \caption{Illustration of \emph{Perturbation-Guided Adversarial Alignment (PGADA)}.}
  \label{fig:representation}
\end{figure*}

According to the Theorem~\ref{thm:err}, the optimal transport can be easily misguided by considering perturbed images. The goal of PGADA is to generate a set of augmented data to derive a more robust embedding model and relieve the perturbation in images. Recently, MaxUp~\cite{gong2021maxup} synthesized augmented data by \textit{minimizing the maximum loss} over the augmented data $x_{p}$, which can be formally written as follows.
\begin{equation}\small
    \min_{\phi,\theta} E_{\{x,y\}\sim \mathcal{D}}[\max_{x_{p}} L(\theta(\phi(x_{p})),  y)],
    \label{eq:maxup}
\end{equation}
and can be easily minimized with stochastic gradient descent (SGD). Specifically, MaxUp samples a batch of augmented data  $x_{p}$ and compute the gradient of the data point which has the highest loss $L$. Therefore, the model would  learn the hardest example over all augmented data $x_{p}$.

However, it's hard to collect sufficient labels for each class in few-shot learning. Therefore, instead of maximizing the empirical risk of the labeled data by Eq. (\ref{eq:maxup}), we introduce a self-supervised learning-based objective.
\begin{equation}\small
    \min_{\phi} E_{\{x\}\sim \mathcal{D}}[\max_{x_{p}} M(\phi(x_{p}),  \phi(x))],
    \label{eq:adversarial}
\end{equation}
where $M$ is the comparison module in few-shot learning, e.g., $l_2$-norm. By maximizing the distance between  $\phi(x_{p})$ and  $\phi(x)$, PGADA generates perturbed image $x_p$, which is less similar to the original image $x$ in embedding space, as a \emph{hard} example. To effectively generate the perturbed data, we introduce a semantic-aware perturbation generator to synthesize the augmented data.
\begin{equation}\small
    x_p = G(x) \emph{ , s.t. } \Vert{x_p - x}\Vert_2^2 \leq \epsilon,
\end{equation}
where $G$ is a model to generate the perturbed image $x_p$.\footnote{We employ a $3$-layer convolutional neural network as our $G$.} Besides, we utilize dropout~\cite{ian2014generative} to provide the randomness of our model. Compared to conventional adversarial training techniques~\cite{samangouei2018defense,wang2018low}, which usually sample the perturbed images from an i.i.d distribution $x_{p} \sim P(\cdot|x)$, e.g., Gaussian distribution, our method can encode the semantic of the image $x$ without requiring many samples to achieve convergence~\cite{gong2021maxup}. 

At the training phase (illustrated in the left of Fig.~\ref{fig:representation}),  we also minimize the empirical risk of the perturbed data $x_p$ to ensure that the generator persists enough information to predict the original label $y$ with KL divergence~\cite{phoo2020self}. The objective is,
\begin{equation}\small
   \max_G  M(\phi(G(x)),  \phi(x)) - KL(\theta(\phi(G(x))), y). 
\label{eq:overall_1}
\end{equation}
Then, we adopt stochastic gradient descent (SGD)~\cite{bottou2012stochastic} to train our generator $G$. It is worth noting that we fix the parameters of the embedding model $\phi$ and $\theta$ when training the generator $G$ to stabilize the training process~\cite{antoniou2017data}.

In addition, we minimize the KL divergence of the original data $x$ and the perturbed data point $x_p$ to train the embedding model (illustrated in the right of Fig.~\ref{fig:representation}).
\begin{equation*}\small
    L_{ori} =  KL(\theta(\phi(x)), y)),
\quad
    L_{adv} = KL(\theta(\phi(x_p)), y)) .
\end{equation*}

To enhance the generalizability of the embeddings, we also leverage the unlabeled data by the auxiliary contrastive self-supervised learning~\cite{chen2020simple}. At each iteration, we sample $N$ images either from training or testing set,\footnote{Note that it is valid to access the images from testing set in few-shot learning,  which is named transductive few-shot learning~\cite{phoo2020self}.} and generate $2$ augmentations for each images, i.e., each augmented image has $1$ positive example and  $2N-1$ corresponding negative examples with the self-supervised loss defined as follows.
\begin{equation}\small
  L_{self} = \frac{1}{2N} \sum_{k=1}^{N}[\ell(2k-1, 2k)+\ell(2k, 2k-1)].
\label{eq:ssl} 
\end{equation}
$\ell$ denotes the \textit{NT-Xent} Loss~\cite{chen2020simple}, which can be written as,
\begin{align*}\small
    \ell(i,j) = -\log \frac{\exp{(cos(z_i,z_j)/ \tau )}}{\sum_{k=1}^{2N}\mathbbm{1}_{k\neq i}\exp{(cos(z_i,z_k)/ \tau)}},
\end{align*}
where $z_*$ is defined as $W\phi(x_*)$, and $W$ is a trainable projection matrix. Summing up, the overall all objective becomes
\begin{equation}\small
  \min_{\phi,\theta} L_{ori} + \lambda_1 L_{adv} + \lambda_2 L_{self},
\label{eq:overall}   
\end{equation}
where $\lambda_1$ and $\lambda_2$ are the trade-off parameters between each loss. Similarly, the classifier $\theta$ is trained by minimizing $L_{ori}$ and $L_{adv}$. The pseudo code is presented in Algorithm~\ref{alg:prada}.
\label{appendix:pseudo}
\begin{algorithm}[t]

    \caption{PGADA}
    \begin{algorithmic}[1]
    \REQUIRE  Dataset $\mathcal{D}$, comparison module $M$, learning rate $\eta$, trade-off parameters $\lambda_1$ and $\lambda_2$.
    \ENSURE Embedding model $\phi$ 
    \STATE Initialize generator $G$, embedding model $\phi$, and classifier $\theta$.
    \FOR{$\{x,y\}$ in $\mathcal{D}$}
        \STATE \textit{\# fixed $\phi, \theta$, update $G$} 
        \STATE $x_p$ = $G(x)$,
        \ $L_{dist}$ = $-M(\phi(x_p),  \phi(x))$,\ 
        $L_{adv}$ = $KL(\theta(\phi(x_p)),y))$
        \STATE {$G \gets G - \eta \nabla(L_{dist} + L_{adv})$ } \textit{\# Generated less similar data points with perturbations.}
        \STATE \textit{\# fixed $G$, update $\phi, \theta$}
        \STATE $x_p$ = $G(x)$,
        \ $L_{ori}$ = $KL(\theta(\phi(x)), y))$, 
        \ $L_{adv}$ = $KL(\theta(\phi(x_p)),y))$
        \STATE $\phi \gets \phi - \eta \nabla$ ($L_{ori} + \lambda_1 L_{adv} + \lambda_2 L_{self})$ 
        \STATE $\theta \gets \theta - \eta \nabla$ ($L_{ori} + \lambda_1 L_{adv} )$
        \textit{\# classifying the generated samples correctly.}
    \ENDFOR
    \end{algorithmic}
    \label{alg:prada}
\end{algorithm}



\subsection{Regularized Optimal Transportation}
After deriving a robust embedding model, we extend the original transportation (Eq.~\ref{eq:basic_ot}) with negative entropy regularization to align the support set and the query set. Thus, the transport plan is penalized as follows.
\begin{align}\small
      \pi^{\ast} = \argmin_{\pi}\sum_{\substack{x_{s,i} \sim \hat{\mu}_{s}\\ x_{q,j} \sim \hat{\mu}_{q}} }\beta w(x_{s,i},x_{q,j}) \pi(x_{s,i},x_{q,j})  
      +(1-\beta) \pi(x_{s,i},x_{q,j}) \log \pi(x_{s,i},x_{q,j}),
\label{eq:smooth_ot_1}
\end{align}
where $\beta$ is the weight parameter to determine the smoothness of the transportation plan. In other words, the data points in the support set take more data points in the query set as anchors for alignment. Accordingly, we can better align the support set to the query set because each labeled data point is representative, especially with limited data points and labels.



\section{Experiment}

We compare PGADA to eleven baselines, including conventional few-shot learning methods and adversarial data augmentation methods on three real datasets. 

\subsection{Experiment Setup}
\label{sec:experiment_setup}

\subsubsection{Datasets.}
Following~\cite{bennequin2021bridging}, we validate our framework on three benchmark datasets in few-shot learning.
1) \textit{CIFAR100}~\cite{krizhevsky2009learning} consists of $60,000$ images, evenly  distributed  in  $100$  classes ($64$ classes for training, $10$ classes for validation, and 25 classes for testing). 2) \textit{miniImageNet}~\cite{triantafillou2019meta} is a subset of ImageNet, with $60,000$ images from $100$ classes ($64$ classes for training, $16$ classes for validation, and $20$ classes for testing). 3) \textit{FEMNIST}~\cite{caldas2018leaf} is a dataset with $805, 263$ handwritten characters in $62$ classes ($42$ classes for training, $10$ classes for validation, and $10$ classes for testing).

\subsubsection{Evaluation.}
Following~\cite{bennequin2021bridging}, the average top-$1$ accuracy scores with $95\%$ confidence interval from $2000$ runs are reported. All experiments are under the $5$-way setting. Note that we conduct the tasks of $1$-shot and $5$-shot with $8$-target and $16$-target, i.e., $1$ or $5$ instances per class in the support set and $8$ or $16$ instances in query set, in CIFAR100, and miniImageNet. While in FEMNIST,  we only adopt tasks of $1$-shot and $1$-target, restricted by the setting of the dataset.
\subsubsection{Implementation.}
We use a $4$-layer convolutional network as the embedding model $\phi$ on CIFAR100 and FEMNIST, and ResNet18 for miniImageNet. As a general framework, we combine PGADA with two classifiers, i.e., ProtoNet and MatchingNet. The hyperparameters are selected by grid search with $\eta = 1e-3$, $b=128$, $d = 128$, $\lambda_1 =1$, $\lambda_2 =1$, $\beta =0.5$, respectively. Note that we also employ transductive batch normalization~\cite{liu2018learning}. Besides, SGD optimizer~\cite{kingma2014adam} is adopted to train all models in $200$ epochs with early stopping. All experiments are implemented in a server with a Intel(R) Core(TM) i9-9820X CPU@3.30GHz, and a GeForce RTX 3090 GPU.
\subsection{Quantitative Analysis}
\label{sec:quantitative}
\begin{table*}[t]\small
\resizebox{\columnwidth}{!}{%
    \centering
    \begin{tabular}{ c|c|c|c|c|c|c|c|c|c }
\hline
  \multirow{3}*{Dataset}  &  \multicolumn{4}{c|}{CIFAR100} &  \multicolumn{4}{c|}{miniImageNet} & 
  FEMNIST\\

    \cline{2-10}
    &
    \multicolumn{2}{c|}{8-target} &
    \multicolumn{2}{c|}{16-target}&
    \multicolumn{2}{c|}{8-target} &
    \multicolumn{2}{c|}{16-target}&
    1-target \\
    \cline{2-10}
    & 1-shot& 5-shot & 
      1-shot& 5-shot & 
      1-shot& 5-shot & 
      1-shot& 5-shot & 
      1-shot \\
    \hline
\multicolumn{10}{c}{Few-shot Learning} \\
  \hline 
   MatchingNet~\cite{vinyals2016matching}& $30.71_{\pm0.38}$ & $41.15_{\pm0.45}$& 
  $31.00_{\pm0.34}$& $41.83_{\pm0.39}$&
  $35.26_{\pm0.50}$ & $44.75_{\pm0.55}$& 
  $37.20_{\pm0.48}$& $44.22_{\pm0.52}$&
  $84.25_{\pm0.71}$   \\
  
  ProtoNet~\cite{snell2017prototypical}& $30.02_{\pm0.40}$& $42.77_{\pm0.47}$& $30.29_{\pm0.33}$& $42.52_{\pm0.41}$&
  $36.37_{\pm0.50}$& $47.58_{\pm0.57}$&
  $35.69_{\pm0.45}$& $46.29_{\pm0.53}$&
  $84.31_{\pm0.73}$ \\

  TransPropNet~\cite{liu2018learning}  & 
  $34.15_{\pm0.39}$  & $47.39_{\pm0.42}$&
  $34.20_{\pm0.40}$ & $44.31_{\pm0.38}$&
  $24.10_{\pm0.27}$& $27.24_{\pm0.33}$&
  $25.38_{\pm0.30}$& $28.05_{\pm0.30}$&
  $86.42_{\pm0.76}$ \\
  
  FTNET~\cite{dhillon2019baseline} & 
  $28.91_{\pm0.37}$ & $37.28_{\pm0.40}$ & $28.66_{\pm0.31}$& $37.37_{\pm0.33}$&
  $39.02_{\pm0.46}$ & $51.27_{\pm0.45}$ & 
  $39.70_{\pm0.40}$ & $52.00_{\pm0.37}$ & 
  $86.13_{\pm0.71}$ \\
  
  TP~\cite{bennequin2021bridging} & 
  $34.00_{\pm0.46}$ & $49.71_{\pm0.47}$ & $35.55_{\pm0.41}$ & $50.24_{\pm0.39}$ &
  $40.49_{\pm0.54}$ & $59.85_{\pm0.49}$ &
  $43.83_{\pm0.51}$& $55.87_{\pm0.42}$&
  $93.63_{\pm0.63}$ \\
\hline
   \multicolumn{10}{c}{Adversarial Data Augmentation} \\
  \hline

  MixUp~\cite{zhang2017mixup} & 
 $37.82_{\pm0.47}$ & 
 $52.57_{\pm0.47}$ & 
 $38.52_{\pm0.42}$ & 
 $53.33_{\pm0.40}$ & 
 $42.98_{\pm0.54}$ & 
 $57.22_{\pm0.48}$ &
 $43.64_{\pm0.48}$ & 
 $57.33_{\pm0.42}$ &
 $97.22_{\pm0.46}$
  \\
  
  CutMix~\cite{yun2019cutmix} & 
 $39.36_{\pm0.48}$ & 
 $54.76_{\pm0.48}$ & 
 $40.05_{\pm0.44}$ & 
 $55.44_{\pm0.40}$ & 
 $35.50_{\pm0.52}$ & 
 $45.50_{\pm0.56}$ &
 $35.78_{\pm0.48}$ & 
 $44.85_{\pm0.52}$ &
 $96.89_{\pm0.49}$
  \\
  
  Autoencoder~\cite{schonfeld2019generalized} & 
  $39.05_{\pm0.50}$ & $53.24_{\pm0.47}$&
  $39.82_{\pm0.44}$ & $53.88_{\pm0.40}$&
  $45.36_{\pm0.56}$ & $57.69_{\pm0.51}$&
  $45.65_{\pm0.52}$ & $57.39_{\pm0.44}$&
  $96.53_{\pm0.43}$
  \\

  AugGAN~\cite{huang2018auggan} &  
  $39.54_{\pm0.50}$& $53.05_{\pm0.47}$&
  $39.50_{\pm0.45}$& $53.42_{\pm0.39}$&
  $44.65_{\pm0.55}$& $57.55_{\pm0.50}$&
  $44.91_{\pm0.49}$& $57.10_{\pm0.42}$&
  $96.42_{\pm0.52}$
  \\

  MaxEntropy~\cite{zhao2020maximum} &  
 $38.14_{\pm0.40}$ & 
 $51.02_{\pm0.56}$ & 
 $38.21_{\pm0.34}$ & 
 $51.33_{\pm0.52}$ & 
 $48.21_{\pm0.36}$ & 
 $57.67_{\pm0.63}$ &
 $48.99_{\pm0.21}$ & 
 $59.01_{\pm0.44}$ &
 $97.19_{\pm0.51}$
 
 \\
 
  MaxUp~\cite{gong2021maxup} &  
  $34.84_{\pm0.44}$& 
  $47.51_{\pm0.46}$&
  $35.20_{\pm0.40}$& 
  $47.63_{\pm0.39}$&
  $37.62_{\pm0.55}$& 
  $48.65_{\pm0.58}$&
  $38.13_{\pm0.50}$& 
  $49.19_{\pm0.51}$&
  $96.48_{\pm0.53}$

  \\

  \hline
   \multicolumn{10}{c}{Ours} \\
  
    \hline

    PGADA (ProtoNet) &  
    $42.16_{\pm 0.52}$ &          $\mathbf{56.52_{\pm0.47}}$ &
    $\mathbf{42.73_{\pm0.46}}$ & $\mathbf{56.83_{\pm0.40}}$ &
    $55.44_{\pm0.61}$ & $\mathbf{67.34_{\pm0.49}}$ &
    $55.69_{\pm0.62}$ & $\mathbf{66.90_{\pm0.50}}$ &
    $\mathbf{97.98_{\pm0.40}}$ \\
    PGADA (MatchingNet) & 
    $\mathbf{42.25_{\pm0.53}}$ &
    $50.98_{\pm0.45}$ &
    $42.60_{\pm0.45}$ &
    $51.80_{\pm0.39}$ &
    $\mathbf{56.15_{\pm0.61}}$ &
    $63.08_{\pm0.49}$ &
    $\mathbf{56.12_{\pm0.57}}$ &
    $63.61_{\pm0.45}$ &
    $97.96_{\pm0.39}$ \\
    \hline
\end{tabular}%
}
\caption{Accuracy comparison of the three datasets with two types of baselines.}
\label{table:quantitative}
\end{table*}


\subsubsection{Few-shot learning.} We first compare five state-of-the-art few-shot learning methods, including  \textit{i) MatchingNet~\cite{vinyals2016matching}}, \textit{ii) ProtoNet~\cite{snell2017prototypical}}, \textit{iii) TransPropNet~\cite{liu2018learning}}, \textit{iv) FTNET~\cite{dhillon2019baseline}},
and \textit{v) Transported Prototypes (TP)~\cite{bennequin2021bridging}}. As shown in Table~\ref{table:quantitative},  PGADA outperforms the best baseline (TP) by at least $13.12\%$ in CIFAR100, $12.51\%$  in miniImageNet, and $4.65\%$ in FEMNIST, respectively. Note that PGADA achieves the best improvement on miniImageNet, since the image size of miniImageNet is the  largest one compared to CIFAR100 and FEMNIST. In addition, PGADA achieves better improvement in the tasks of 1-shot  (i.e., $22.46\%$) than the tasks of $5$-shot ($14.77\%$) averagely, which demonstrates the robustness of our method, especially with limited labeled data. Compared with the first four baselines, i.e., ProtoNet, MatchingNet, TransPropNet, and FTNET, our method achieves consistent improvement, $41.48\%$, $39.78\%$, $71.61\%$, and $54.85\%$, respectively,  since these methods do not consider the inherent distribution shift between the support and query set. While TP also utilizes optimal transport to align the support and query set, it still shows relatively weak performance compared to PGADA because the transportation plan of TP is misguided by the small perturbations in the images, as proved in Theorem~\ref{thm:err}. Since PGADA is a model agnostic adversarial alignment framework, we equip PGADA with different classifiers, e.g., ProtoNet and MatchingNet. Our framework outperforms the baseline by $39.78\%$ and $21.81\%$ in ProtoNet and MatchingNet, respectively, manifesting the generability of PGADA. We observe that ProtoNet outperforms MatchingNet in the 5-shot case as the prototypes reduce the bias by averaging the embedding vectors, which is more robust. 

\subsubsection{Adversarial data augmentation.}
In addition, as our method is closed to adversarial data augmentation, we also compares six adversarial data augmentation methods, including \textit{vi) MixUp} \cite{zhang2017mixup}, \textit{vii) CutMix} \cite{yun2019cutmix}, \textit{viii)  Autoencoder} \cite{schonfeld2019generalized},  \textit{ix) AugGAN} \cite{huang2018auggan}, \textit{x) MaxEntropy} \cite{zhao2020maximum}, and \textit{xi) MaxUp} \cite{gong2021maxup}. According to Table~\ref{table:quantitative}, PGADA outperforms MixUp and CutMix by $9.12\%$ and $8.84\%$ on average. Compared with the Autoencoder and AugGAN, our method improved by $11.77\%$ and $12.41\%$ because they synthesize similar patterns to the original images to alleviate the data sparsity problem, which cannot create new information that is not included in the given data~\cite{gong2021maxup}.  In contrast, PGADA is able to explore the perturbed data that is most likely to confuse the model, which can be regarded as hard examples. Compared to MaxEntropy, which uses opposite gradients to search the worst data point, we observe that PGADA outperforms MaxEntropy by $6.52\%$ on average since PGADA adds noise to the generator, increasing the uncertainty of the generated data. Then, by adding these data points to the training phase, the model can defense against more unknown perturbations attacking, resulting in higher accuracy. In addition, our method also significantly outperforms MaxUp because the selected worst data point in MaxUp is still near the original images, resulting in less robustness. In addition, PGADA is more suitable for few-shot learning, as it generates the perturbations in a self-supervised manner by comparing the embedding of original images and perturbed images to explore hard examples.

\begin{table*}[t]\small
\resizebox{\columnwidth}{!}{%
    \centering
    \begin{tabular}{ c|c|c|c|c|c|c|c|c|c }
\hline
  \multirow{3}*{Dataset}  &  \multicolumn{4}{c|}{CIFAR100} &  \multicolumn{4}{c|}{miniImageNet} & 
  FEMNIST\\

    \cline{2-10}
    &
    \multicolumn{2}{c|}{8-target} &
    \multicolumn{2}{c|}{16-target}&
    \multicolumn{2}{c|}{8-target} &
    \multicolumn{2}{c|}{16-target}&
    1-target \\
    \cline{2-10}
    & 1-shot& 5-shot & 
      1-shot& 5-shot & 
      1-shot& 5-shot & 
      1-shot& 5-shot & 
      1-shot \\
  \hline
    
  PGADA &  
 $\mathbf{42.16_{\pm 0.52}}$ &          $\mathbf{56.52_{\pm0.47}}$ &
 $\mathbf{42.73_{\pm 0.46}}$ & $\mathbf{56.83_{\pm0.40}}$ &
 $\mathbf{55.44_{\pm0.61}}$ &  $\mathbf{67.34_{\pm0.49}}$ &
 $\mathbf{55.69_{\pm0.62}}$ &  $\mathbf{66.90_{\pm0.50}}$ &
 $\mathbf{97.98_{\pm0.40}}$ \\
 \hline

  \multicolumn{10}{c}{Generator} 
  \\
  \hline
    
  fixed G & 
  $38.58_{\pm0.48}$  & $52.41_{\pm0.47}$&
  $39.26_{\pm0.43}$  & $52.67_{\pm0.39}$&
  $43.50_{\pm0.55}$& $55.65_{\pm0.50}$&
  $43.48_{\pm0.51}$& $55.42_{\pm0.43}$&  
  $96.41_{\pm0.52}$  \\
  
  w/o noise & 
  $37.16_{\pm0.47}$& $50.12_{\pm0.46}$&
  $37.73_{\pm0.41}$& $50.50_{\pm0.38}$& 
  $44.06_{\pm0.56}$& $56.97_{\pm0.48}$&
  $44.42_{\pm0.49}$& $56.96_{\pm0.42}$&
  $96.89_{\pm0.48}$ \\

  w/o KL& 
  $37.30_{\pm0.47}$ & $50.79_{\pm0.46}$&
  $37.91_{\pm0.42}$ & $51.35_{\pm0.39}$&
  $44.22_{\pm0.54}$& $55.04_{\pm0.49}$&
  $44.21_{\pm0.49}$& $53.96_{\pm0.41}$&
  $96.49_{\pm0.48}$ \\

  \hline
  \multicolumn{10}{c}{Regularized Optimal Transport (OT)}
  \\
  \hline 
   w/o OT & 
  $35.76_{\pm0.41}$ &
  $54.06_{\pm0.45}$ & 
  $35.66_{\pm0.35}$ & 
  $54.09_{\pm0.38}$ &
  $44.30_{\pm0.52}$ & 
  $61.23_{\pm0.53}$ &
  $44.15_{\pm0.46}$ & 
  $60.86_{\pm0.48}$ &
  $94.03_{\pm0.48}$ \\
  
    TP~\cite{bennequin2021bridging} & 
    $34.00_{\pm0.46}$ & 
    $49.71_{\pm0.47}$ & 
    $35.55_{\pm0.41}$ & 
    $50.24_{\pm0.39}$ &
    $40.49_{\pm0.54}$ & 
    $59.85_{\pm0.49}$ &
    $43.83_{\pm0.51}$ & 
    $55.87_{\pm0.42}$ &
    $93.63_{\pm0.63}$ \\

    TP  w/o OT &  
    $33.07_{\pm0.38}$ &          
    $50.99_{\pm0.44}$ &
    $32.96_{\pm0.32}$ & 
    $50.71_{\pm0.37}$ &
    $38.07_{\pm0.45}$ &  
    $55.31_{\pm0.51}$ &
    $37.94_{\pm0.41}$ &  
    $55.11_{\pm0.44}$ &
    $91.84_{\pm0.56}$ \\
   \hline
  \multicolumn{10}{c}{Self-supervised Learning (SSL)} 
  \\
   \hline
    w/o SSL &   
    $39.33_{\pm0.50}$  & $53.66_{\pm0.47}$&
    $40.31_{\pm 0.44}$ & $54.23_{\pm 0.40}$ &
    $47.96_{\pm0.57}$& $61.38_{\pm0.49}$&
    $48.70_{\pm 0.52}$ & $61.44_{\pm 0.43}$ &
    $97.07_{\pm0.48}$ \\
    
    \hline
\end{tabular}%
}
\caption{The results of ablation studies.}

\label{table:ablation}
\end{table*}
\subsection{Ablation Studies}
\label{sec:ablation}
We conduct ablation studies to evaluate the importance of different modules in PGADA. Note that we only present the results of PGADA with ProtoNet as PGADA with MatchingNet shows similar ones.

\subsubsection{Effect of Generator.}
We compare PGADA with three different variants of the generator, including i) \textit{fixed $G$}, fixing the parameters of the generator, ii) \textit{w/o noise}, removing the noise function of the generator, and iii) \textit{w/o KL}, removing the classification loss of perturbed data. When we fix the parameters of the generator, we can observe that the performance of PGADA drops by $8.46\%$, $24.31\%$, and $1.63\%$ in three datasets, respectively. It demonstrates that our trainable generator is able to extract the information from original images $x$ to generate meaningful perturbed images $x_p$. As we remove the noise, it also shows a consistent decline in performance, which demonstrates that leveraging some randomness in the training process is helpful to explore the perturbation. Lastly, our method without classification loss (i.e., \textit{w/o KL}) still outperforms TP~\cite{bennequin2021bridging}. It demonstrates that the self-supervised-learning-based objective function (Eq. \ref{eq:adversarial}) works well to learn the inherent information from the original data, leading to a robust embedding model. However, the classification loss of the perturbed data also boosts the model capability 
as it regularizes the generator by preserving useful information to predict a correct label, rather than exploration via random walk, which is less efficient. 

\subsubsection{Effect of Regularized Optimal Transportation.} 
Here, we investigate the effect of regularized optimal transportation (OT), which plays a crucial role during the evaluation phase in few-shot learning under the support-query shift. It shows that OT significantly improves the model's capability since it aligns the distributions of the support and query set. Even though the performance of PGADA drops as we remove OT. PGADA still outperforms the state-of-the-art method TP \cite{bennequin2021bridging}, which also utilizes OT by $4.86\%$. Also, compared TP to TP w/o OT, we observe that optimal transportation in TP does not perform well as in PGADA. This observation echoes the motivation of this work (Theorem~\ref{thm:err}), i.e., the small perturbations misguide the optimal transportation plan. Also, this result manifests that a robust embedding model indeed alleviates the  support-query shift in few-shot learning.  

\subsubsection{Effect of Self-Supervised Learning.} 
Last, we evaluate the effect of self-supervised learning. Equipped with the contrastive loss, PGADA improves by $5.83\%$, $12.13\%$, and $0.94\%$ in three datasets, respectively, since the model leverages the structural information of the unlabeled data in the training phase and thus leaps in model performance. The results also demonstrate that PGADA and self-supervised learning can incorporate together to explore the information from the training set and testing set to improve the generality of the embedding model.  Understanding why this combination performs well is interesting for future works.

\section{Related Work}
\label{sec:related}
\subsubsection{Few-shot Learning.} 
Few-shot learning can be divided into three main categories, including optimization-based method~\cite{finn2017model}, hallucination-based method~\cite{hariharan2017low}, and metric-based method~\cite{vinyals2016matching}. On the one hand, the optimization-based methods aim to train a quickly adaptive model and quickly adapt to other tasks by fine-tuning~\cite{finn2017model,boudiaf2020transductive}. On the other hand, the hallucination-based methods hallucinate the scarce labeled data by synthesizing representations~\cite{hariharan2017low} with Generative Adversarial Networks (GANs)~\cite{antoniou2017data}. Our work is most relevant to the metric-based methods where Vinyals et al.~\cite{vinyals2016matching} and Snell et al.~\cite{snell2017prototypical}  introduce the pairwise (and classwise) metrics to determine the label of the query set according to the support set.  Sung et al.~\cite{sung2018learning} model the non-linear relation between class representations and queries by neural networks. Moreover, cross-domain few-shot learning~\cite{chen2019closer} argues that the training and testing sets may not lie in the same distribution. Thus, several techniques such as optimal transport~\cite{courty2016optimal}, self-supervised learning~\cite{zhao2021domain,phoo2020self} have been introduced to relieve the domain shift. 
While the domain shift not only occurs between the training set and the testing set but also between the support set and the query set~\cite{bennequin2021bridging}. In addition, the optimal transport plan can be easily misguided by the perturbation in the images and leads to unacceptable performance.

\subsubsection{Data Augmentation and Adversarial Training.}
Data augmentation has been widely used in machine learning with various transformations on a single image ~\cite{simonyan2014very}, e.g., resizing, flipping, rotation, cropping, or multiple images, e.g., MixUp~\cite{zhang2017mixup} and CutMix~\cite{yun2019cutmix}. Another line of studies~\cite{chen2020simple,xie2020self} works on several self-training schemes by maximizing the pairwise similarity of augmented data. However, the aforementioned methods cannot create new information not included in the given data~\cite{theagarajan2019shieldnets} since they synthesize and train on similar images. In contrast, adversarial training~\cite{samangouei2018defense,zhao2020maximum} has been developed to defend against adversarial attacks, which conducts the attack to generate perturbed examples from clean data that the model misclassifies. Then, the generated examples are used as training data to compensate for the model weaknesses, contributing to model robustness. A similar work, MaxUP~\cite{gong2021maxup}, generates a set of random augmented data and searches the hardest example that would maximize the classification loss. However, MaxUp requires numerous random augmented data to explore more information, thus leading to ineffectiveness. In contrast, PGADA directly generates hard samples in a self-supervised manner, which is simpler and computationally efficient.

\section{Conclusion}
\label{sec:conclusion}

In this paper, we propose \emph{Perturbation-Guided Adversarial Alignment (PGADA)} to solve the support-query shift in few-shot learning. Our key idea is to generate perturbed images that are hard to classify and then train on these perturbed data to derive a more robust embedding model and alleviate the misestimation of optimal transportation. In addition, a negative entropy regularization is introduced to obtain a smooth transportation plan. The experiment results manifest that PGADA outperforms eleven baselines by at least $13.12\%$ in CIFAR100, $12.51\%$  in miniImageNet, and $4.65\%$ in FEMNIST, respectively. Future works include applying PGADA to other computer vision tasks and incorporating it with other data augmentation schemes.

\section*{Acknowledgement}
S. Jiang is supported by the science and
technology plan project in Huizhou (No. 2020SD0402030) 

\bibliographystyle{splncs04}
\bibliography{main}

\begin{thebibliography}{10}
\providecommand{\url}[1]{\texttt{#1}}
\providecommand{\urlprefix}{URL }
\providecommand{\doi}[1]{https://doi.org/#1}

\bibitem{antoniou2017data}
Antoniou, A., Storkey, A., Edwards, H.: Data augmentation generative
  adversarial networks. In: ICLR (2017)

\bibitem{bennequin2021bridging}
Bennequin, E., Bouvier, V., Tami, M., Toubhans, A., Hudelot, C.: Bridging
  few-shot learning and adaptation: New challenges of support-query shift.
  ECML-PKDD  (2021)

\bibitem{bottou2012stochastic}
Bottou, L.: Stochastic gradient descent tricks. In: Neural networks: Tricks of
  the trade, pp. 421--436. Springer (2012)

\bibitem{boudiaf2020transductive}
Boudiaf, M., Masud, Z.I., Rony, J., Dolz, J., Piantanida, P., Ayed, I.B.:
  Transductive information maximization for few-shot learning. arXiv preprint
  arXiv:2008.11297  (2020)

\bibitem{caldas2018leaf}
Caldas, S., Duddu, S.M.K., Wu, P., Li, T., Kone{\v{c}}n{\`y}, J., McMahan,
  H.B., Smith, V., Talwalkar, A.: Leaf: A benchmark for federated settings.
  NeurIPS  (2019)

\bibitem{chen2020simple}
Chen, T., Kornblith, S., Norouzi, M., Hinton, G.: A simple framework for
  contrastive learning of visual representations. In: ICML. pp. 1597--1607.
  PMLR (2020)

\bibitem{chen2019closer}
Chen, W.Y., Liu, Y.C., Kira, Z., Wang, Y.C.F., Huang, J.B.: A closer look at
  few-shot classification. arXiv preprint arXiv:1904.04232  (2019)

\bibitem{courty2016optimal}
Courty, N., Flamary, R., Tuia, D., Rakotomamonjy, A.: Optimal transport for
  domain adaptation. IEEE TPAMI  (2016)

\bibitem{cuturi2013sinkhorn}
Cuturi, M.: Sinkhorn distances: Lightspeed computation of optimal transport.
  NeurIPS  \textbf{26},  2292--2300 (2013)

\bibitem{dhillon2019baseline}
Dhillon, G.S., Chaudhari, P., Ravichandran, A., Soatto, S.: A baseline for
  few-shot image classification. In: ICLR (2019)

\bibitem{finn2017model}
Finn, C., Abbeel, P., Levine, S.: Model-agnostic meta-learning for fast
  adaptation of deep networks. In: ICML (2017)

\bibitem{garcia2017few}
Garcia, V., Bruna, J.: Few-shot learning with graph neural networks. In: ICLR
  (2017)

\bibitem{gong2021maxup}
Gong, C., Ren, T., Ye, M., Liu, Q.: Maxup: Lightweight adversarial training
  with data augmentation improves neural network training. In: CVPR. pp.
  2474--2483 (2021)

\bibitem{ian2014generative}
Goodfellow, I., Pouget-Abadie, J., Mirza, M., Xu, B., Warde-Farley, D., Ozair,
  S., Courville, A., Bengio, Y.: Generative adversarial nets. NeurIPS
  \textbf{27},  2672–2680 (2014)

\bibitem{hariharan2017low}
Hariharan, B., Girshick, R.: Low-shot visual recognition by shrinking and
  hallucinating features. In: CVPR. pp. 3018--3027 (2017)

\bibitem{huang2018auggan}
Huang, S.W., Lin, C.T., Chen, S.P., Wu, Y.Y., Hsu, P.H., Lai, S.H.: Auggan:
  Cross domain adaptation with gan-based data augmentation. In: ECCV. pp.
  718--731 (2018)

\bibitem{jiang2021dataflow}
Jiang, S., Chen, H.W., Chen, M.S.: Dataflow systolic array implementations of
  exploring dual-triangular structure in qr decomposition using high-level
  synthesis. In: ICFPT (2021)

\bibitem{jiang2019fund}
Jiang, S., Yao, X., Long, Q., Chen, J., Jiang, H.: Fund investment decision in
  support vector classification based on information entropy. Review of
  Economics \& Finance  (2019)

\bibitem{kingma2014adam}
Kingma, D.P., Ba, J.: Adam: A method for stochastic optimization. In: ICLR
  (2014)

\bibitem{krizhevsky2009learning}
Krizhevsky, A., Hinton, G., et~al.: Learning multiple layers of features from
  tiny images  (2009)

\bibitem{liu2018learning}
Liu, Y., Lee, J., Park, M., Kim, S., Yang, E., Hwang, S.J., Yang, Y.: Learning
  to propagate labels: Transductive propagation network for few-shot learning.
  In: ICLR (2018)

\bibitem{phoo2020self}
Phoo, C.P., Hariharan, B.: Self-training for few-shot transfer across extreme
  task differences. In: ICLR (2020)

\bibitem{samangouei2018defense}
Samangouei, P., Kabkab, M., Chellappa, R.: Defense-gan: Protecting classifiers
  against adversarial attacks using generative models. In: ICLR (2018)

\bibitem{schonfeld2019generalized}
Schonfeld, E., Ebrahimi, S., Sinha, S., Darrell, T., Akata, Z.: Generalized
  zero-and few-shot learning via aligned variational autoencoders. In: CVPR.
  pp. 8247--8255 (2019)

\bibitem{simonyan2014very}
Simonyan, K., Zisserman, A.: Very deep convolutional networks for large-scale
  image recognition. In: ICLR (2015)

\bibitem{snell2017prototypical}
Snell, J., Swersky, K., Zemel, R.S.: Prototypical networks for few-shot
  learning. In: NeurIPS (2017)

\bibitem{sung2018learning}
Sung, F., Yang, Y., Zhang, L., Xiang, T., Torr, P.H., Hospedales, T.M.:
  Learning to compare: Relation network for few-shot learning. In: CVPR. pp.
  1199--1208 (2018)

\bibitem{theagarajan2019shieldnets}
Theagarajan, R., Chen, M., Bhanu, B., Zhang, J.: Shieldnets: Defending against
  adversarial attacks using probabilistic adversarial robustness. In: CVPR. pp.
  6988--6996 (2019)

\bibitem{triantafillou2019meta}
Triantafillou, E., Zhu, T., Dumoulin, V., Lamblin, P., Evci, U., Xu, K.,
  Goroshin, R., Gelada, C., Swersky, K., Manzagol, P.A., et~al.: Meta-dataset:
  A dataset of datasets for learning to learn from few examples. arXiv preprint
  arXiv:1903.03096  (2019)

\bibitem{vinyals2016matching}
Vinyals, O., Blundell, C., Lillicrap, T., Wierstra, D., et~al.: Matching
  networks for one shot learning. In: NeurIPS (2016)

\bibitem{wang2018low}
Wang, Y.X., Girshick, R., Hebert, M., Hariharan, B.: Low-shot learning from
  imaginary data. In: CVPR. pp. 7278--7286 (2018)

\bibitem{xie2020self}
Xie, Q., Luong, M.T., Hovy, E., Le, Q.V.: Self-training with noisy student
  improves imagenet classification. In: CVPR. pp. 10687--10698 (2020)

\bibitem{yun2019cutmix}
Yun, S., Han, D., Oh, S.J., Chun, S., Choe, J., Yoo, Y.: Cutmix: Regularization
  strategy to train strong classifiers with localizable features. In: ICCV. pp.
  6023--6032 (2019)

\bibitem{zhang2017mixup}
Zhang, H., Cisse, M., Dauphin, Y.N., Lopez-Paz, D.: mixup: Beyond empirical
  risk minimization. In: ICLR (2017)

\bibitem{zhao2021domain}
Zhao, A., Ding, M., Lu, Z., Xiang, T., Niu, Y., Guan, J., Wen, J.R.:
  Domain-adaptive few-shot learning. In: WACV. pp. 1390--1399 (2021)

\bibitem{zhao2020maximum}
Zhao, L., Liu, T., Peng, X., Metaxas, D.: Maximum-entropy adversarial data
  augmentation for improved generalization and robustness. arXiv preprint
  arXiv:2010.08001  (2020)

\end{thebibliography}


\end{document}